\documentclass{article}

\usepackage[preprint, nonatbib]{icbinb2021}

\usepackage[utf8]{inputenc} 
\usepackage[T1]{fontenc}    
\usepackage{hyperref}       
\usepackage{url}            
\usepackage{booktabs}       
\usepackage{amsfonts}       
\usepackage{nicefrac}       
\usepackage{microtype}      
\usepackage{xcolor}         
\usepackage{graphicx}
\usepackage{subcaption}

\title{Is the Number of Trainable Parameters All That Actually Matters?}

\author{%
  Am\'elie Chatelain$^1$ \And Amine Djeghri$^{3}$\thanks{Work done while at LightOn.} \And
  Daniel Hesslow$^1$ \And
  Julien Launay$^{1, 2}$ \And Iacopo Poli$^1$\vspace{0.2cm}\\
  $^1$LightOn \hspace{0.5cm} $^2$LPENS, École Normale Supérieure \hspace{0.5cm} $^3$Sorbonne Universit\'e \vspace{0.2cm}\\
  \texttt{\{firstname\}@lighton.ai}
}

\begin{document}

\maketitle

\begin{abstract}
Recent work has identified simple empirical scaling laws for language models, linking compute budget, dataset size, model size, and autoregressive modeling loss. The validity of these simple power laws across orders of magnitude in model scale provides compelling evidence that larger models are also more capable models. However, scaling up models under the constraints of hardware and infrastructure is no easy feat, and rapidly becomes a hard and expensive engineering problem. We investigate ways to tentatively \emph{cheat} scaling laws, and train larger models for cheaper. We emulate an increase in effective parameters, using efficient approximations: either by \emph{doping} the models with frozen random parameters, or by using fast structured transforms in place of dense linear layers. We find that the scaling relationship between test loss and compute depends only on the \emph{actual} number of trainable parameters; scaling laws cannot be deceived by spurious parameters.
\end{abstract}

\section{Introduction}
Predictably linking model and dataset size with generalization error is a long-standing open question. Discrepancies between classical bias-variance trade-off models and modern practices have been identified \cite{belkin2019reconciling}, with phenomena such as deep double descent \cite{nakkiran2019deep} providing glimpses into a deeper understanding. However, actionable insights for machine learning practitioners have remained elusive.

Recently, simple and general empirical scaling laws for deep learning models have been uncovered. Starting with first relationships for modern convolutional networks \cite{hestness2017deep}, guidelines informing optimal architectures--such as EfficientNet \cite{tan2019efficientnet}--have been derived. Importantly, these laws can extrapolate model performance across scale, motivating the training of increasingly large and capable models \cite{rosenfeld2019constructive}. 

We focus on seminal work on large generative language models \cite{ScalingForNeural}, establishing and using scaling laws to predict the computational requirements for a specific performance level--as pioneered by \cite{hestness2019beyond}. This work showed that the performance of an auto-regressive language model follows a remarkably simple relationship, linking model and dataset size, compute budget, and modeling loss across many orders of magnitude in scale. This make it possible to answer questions central to the efficient training of extreme-scale models, such as: \textit{(1) which model size achieves the best loss given a fixed compute budget}; or \textit{(2) how many samples are necessary to train a model of a given size optimally}.

Unfortunately, these empirical laws also show that training models much larger than current ones comes at a prohibitive cost. State-of-the-art language models such as GPT-3 have required several thousand PF-days to train \cite{gpt3}, and model size has plateaued since then \cite{zeng2021pangu, kim2021changes}. Proposed distributed sharding methods for the training of such extreme-scale models only bridge the memory gap \cite{zero-inf}, and do not fundamentally change the required compute budget. Pending drastic hardware improvements, it is unlikely we will see a 100-1,000x increase in model size as quickly as we have seen in the past.

Accordingly, to enable sustained growth in model size and capabilities, we attempt to \emph{cheat} scaling laws. To do so, we propose to fudge the effective number of parameters in the model. Instead of using only expensive fully trainable parameters, we explore two classes of efficient parameters: either by \emph{doping} the models with cheaper and simpler frozen parameters, or by implementing \emph{structured} transforms in place of dense linear layers. We study these additions under the guise of scaling laws. 

\paragraph{Frozen parameters}
Transformer models with fixed random parameters have previously been studied in Reservoir Transformers \cite{Shen2020ReservoirT}, where some layers of a RoBERTa model \cite{liu2019roberta} are initialized randomly and then never updated. Using an \textit{area under convergence curve} metric, they find this approach to lead to more efficient training. This is motivated by a broader line of work around random projections in machine learning, with applications in random kernel methods \cite{rahimi2007random}, unsupervised feature learning \cite{saxe2011random}, and reservoir computing \cite{dong2020reservoir}. In a language model, frozen parameters are one third cheaper than fully trainable ones: assuming the backward pass is normally twice the cost in FLOP of the forward (backpropagation + update) \cite{ScalingForNeural}, the update step is skipped for frozen parameters. Furthermore, multiplication with random matrices can be offloaded to specialized co-processors \cite{opu}. 

\paragraph{Structured parameters} 
Structured efficient linear layer parametrizations have been proposed to accelerate fully-connected layers \cite{acdc,yang2015deep,sells}. They leverage transforms with an efficient $\mathcal{O}(n \log n)$ implementation, such as the Hadamard Transform, Discrete Fourier Transform or the Discrete Cosine Transform. These structured layers have far fewer trainable parameters (on the order of $n$) but still emulate an operation equivalent to $n^2$ parameters. We initially hypothesize that the scaling laws will depend on the the emulated number of parameters $n^2$ rather than $n$, reducing the memory needs and compute budget for training the model, while maintaining identical modelling loss.

\paragraph{Contributions} We study random and structured parameters in large auto-regressive language models, and show that scaling laws \textbf{cannot} be cheated using spurious parameters. More specifically, we find that in both cases, the empirical scaling laws obtained depend only on the \emph{real} number of trainable parameters. Neither the frozen or structured parameters make pre-training more efficient. 

\section{Methods}

All experiments are based on the training on a language modeling task of a decoder-only auto-regressive Transformer based on the GPT architecture \cite{gpt2}. Our only significant deviation from the architecture is the use of rotary embeddings \cite{rotary}, which enable increased performance. We performed initial pathfinding experiments without rotary embeddings and obtained similar results. Architecture details for the four model sizes considered are available in the supplementary. 

The model is trained on French data obtained using a Common Crawl dump filtered by CCNet \cite{wenzek2019ccnet}. We  use byte-level  byte-pair  encoding, with a vocabulary of 50,262 tokens. Our dataset contains 32 billion tokens, and is similar to the setup of FlauBERT \cite{le2019flaubert}. All validation and test losses are obtained on the \texttt{fr-wiki} dataset, made of 0.5 billion tokens obtained from a dump of French Wikipedia. 

All experiments were run in a distributed setup with four NVIDIA V100-16GB GPUs per node, on the Jean Zay supercomputer. We use a sharded data parallel strategy for training the larger models, similar to ZeRO-3 \cite{rajbhandari2020zero}, and implemented with FairScale \cite{FairScale2021}. Structured transforms are implemented as optimized CUDA kernels and compiled in PyTorch extensions for best performance.

\begin{figure}
    \centering
    \includegraphics[width=.45\textwidth]{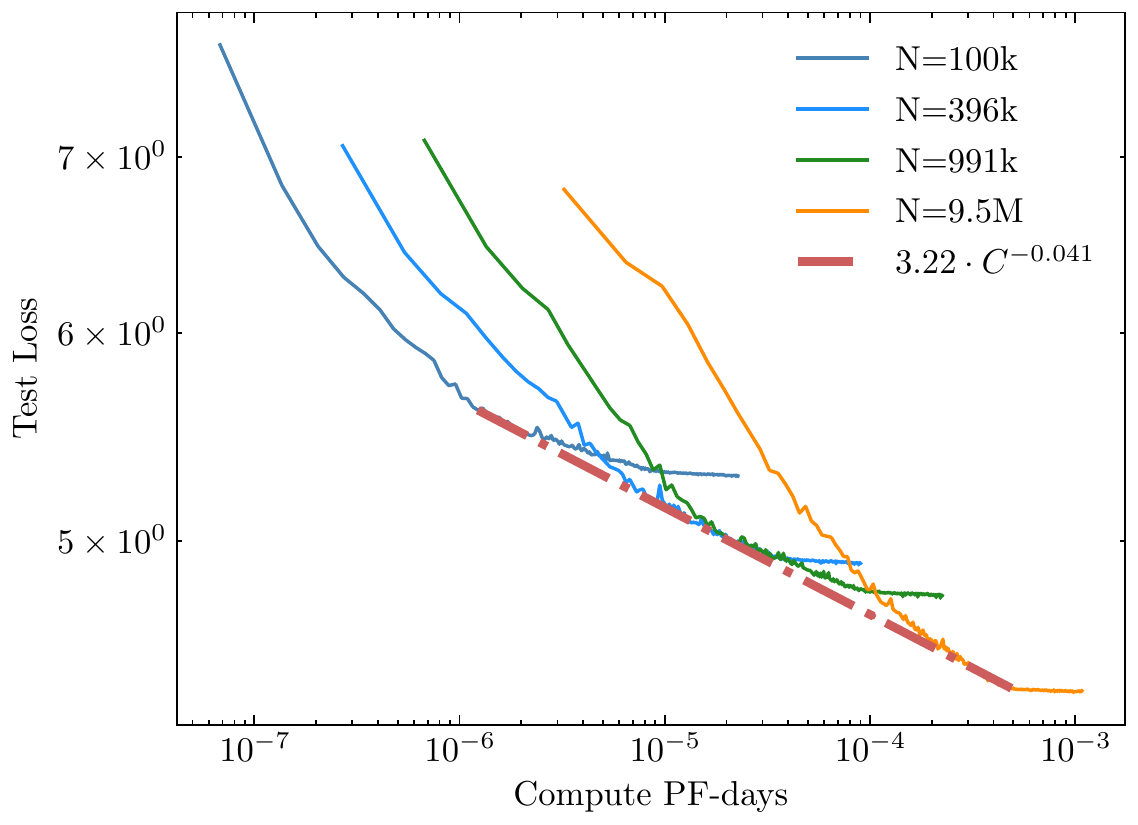}
    \caption{Loss against compute for different model sizes, with the empirical scaling law fit. We highlight that we find an exponent of $-0.041$, which is remarkably close to the $-0.048$ of \cite{henighan2020scaling}, considering that it is obtained on a different dataset in a different language, with a different codebase.
}
    \label{fig:scalinglaw}
\end{figure}

We estimate the compute budget in PF-day using the formula $C=6NBS$ \cite{ScalingForNeural}, where $N$ is the number of trainable non-embedding parameters, $B$ is the number of tokens per batch, $S$ is the number of parameter updates, and $6$ is the factor to account for the forward and backward passes. We verified our experimental setup, and in particular the choice of a French dataset in place of an English one, by reproducing the scaling law fit of \cite{ScalingForNeural} and obtaining a similar exponent -- as shown in Figure \ref{fig:scalinglaw}.

\section{Doping with Frozen Random Parameters}

\begin{figure}
\centering
\begin{minipage}[t]{0.425\textwidth}
\includegraphics[width=\linewidth]{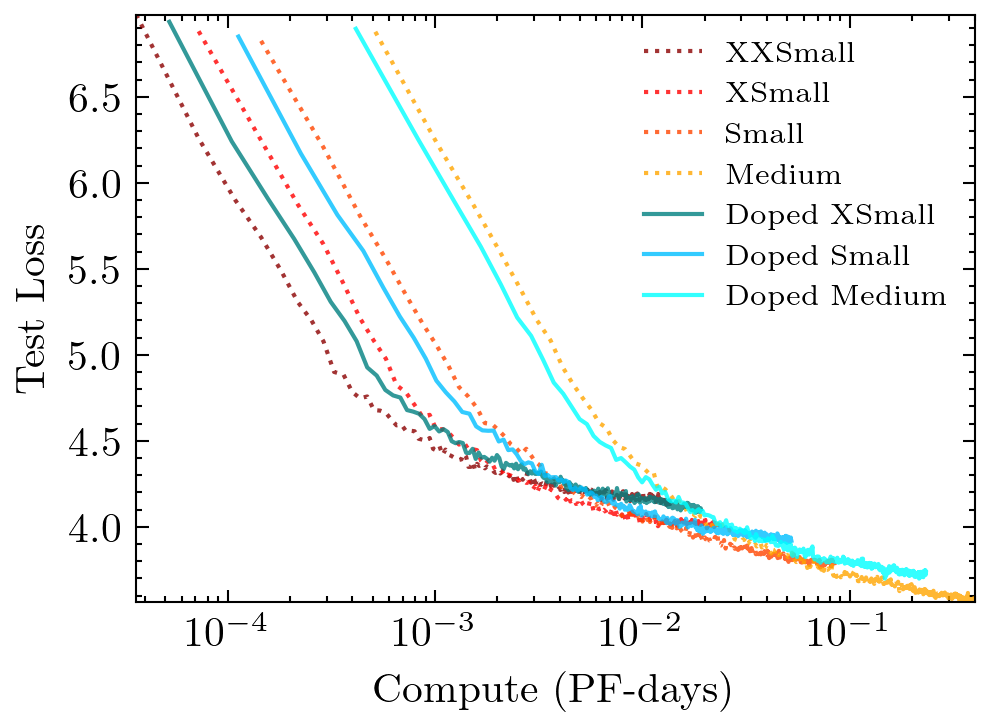}
  \caption{Loss against compute for random parameters costing two thirds as much as trainable ones, as in our implementation.}
  \label{fig:randomdopingspeed1}
\end{minipage}
\hspace{1cm}
\begin{minipage}[t]{0.425\textwidth}
    \centering
    \includegraphics[width=\linewidth]{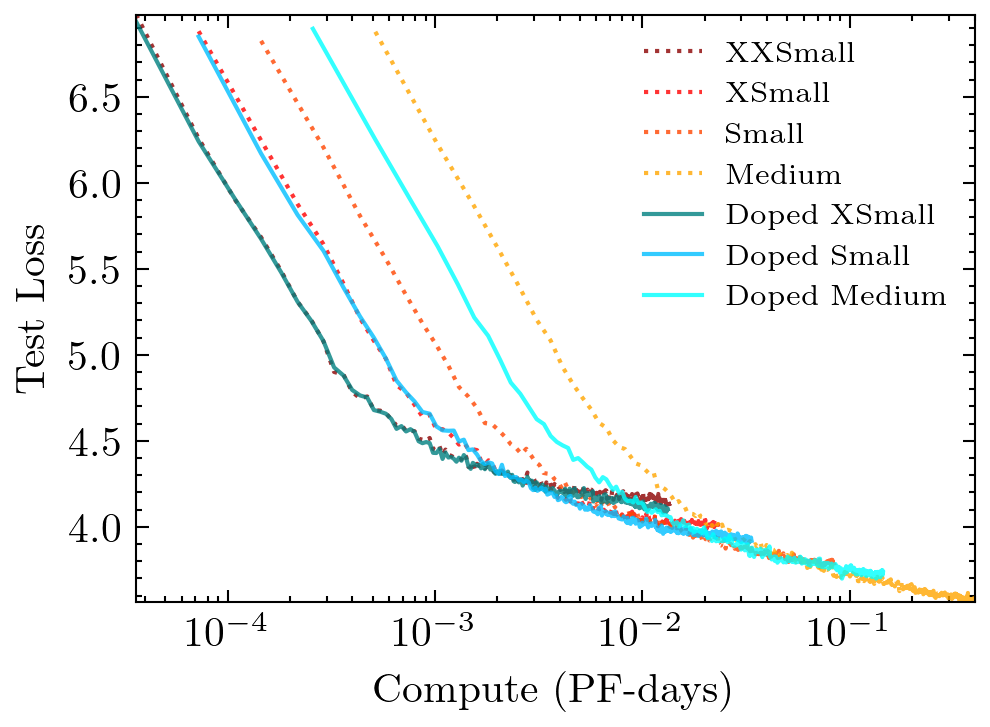}
    \caption{Loss against compute for an ideal random parameter cost of zero, achievable with offloading to dedicated hardware.}
    \label{fig:randomdopingspeedinf}
\end{minipage}
\end{figure}

\subsection{Methods}
Similar to \cite{Shen2020ReservoirT} we replace the decoder layers--consisting of self-attention followed by a multilayer perceptron (MLP)--with a single frozen MLP. In order to approximately preserve the same total number of parameters, we increase the hidden dimension of the feedforward network from $4 \times d_\mathrm{model}$ to $6 \times d_\mathrm{model}$.
By training models with fixed parameters over multiple orders of magnitude, we establish the scaling laws both for the models with fixed random parameters and the baseline models.

\subsection{Results}
The cost of the random parameters depends on the choice of hardware, as well as any potential algorithm used to approximate them. To be able to compare the models with frozen random parameter with the baseline models, we plot the scaling laws for different costs of the random parameters. 

Figure \ref{fig:randomdopingspeed1} shows scaling laws for random parameters costing two thirds as much as trainable parameters, which corresponds to skipping the parameter update step, and which is effectively achieved by our implementation. The baseline models (dotted lines) significantly outperform the models with random parameters (solid lines). Figure \ref{fig:randomdopingspeedinf} shows the scaling behaviour assuming the best-case scenario of free random parameters: doped models reproduce the training curves of a model with the same amount of trainable parameters and no frozen parameters. In other words, the frozen parameters have no effect at all on the loss-compute curve of these models, and are effectively useless for pre-training.

\section{Structured Efficient Linear Layers}
Since frozen layers do not contribute to the loss-compute curve, we turn our attention to structured layers. Structured layers are a natural midpoint between fully frozen and fully trainable layers, implementing a smaller number of \textit{structured} parameters, typically only linearly increasing with layer width, whereas fully-connected layers will increase quadratically. Importantly, and in contrast with sparse layers, structured layers can be efficiently implemented in modern hardware.

\subsection{Methods}
\begin{figure}
\centering
\begin{minipage}[t]{0.425\textwidth}
  \includegraphics[width=\linewidth]{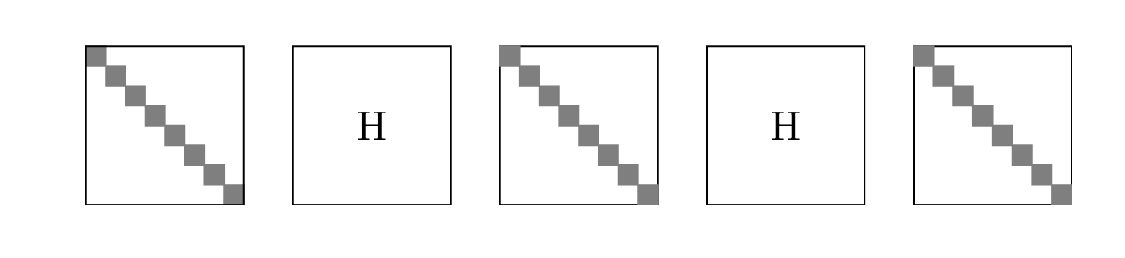} 
  \caption{Adaptive FastFood replaces a linear layer by three learnable diagonal matrices and two Hadamard transforms.}
  \label{fig:fastfood}
\end{minipage}
\hspace{1cm}
\begin{minipage}[t]{0.425\textwidth}
    \includegraphics[width=\linewidth]{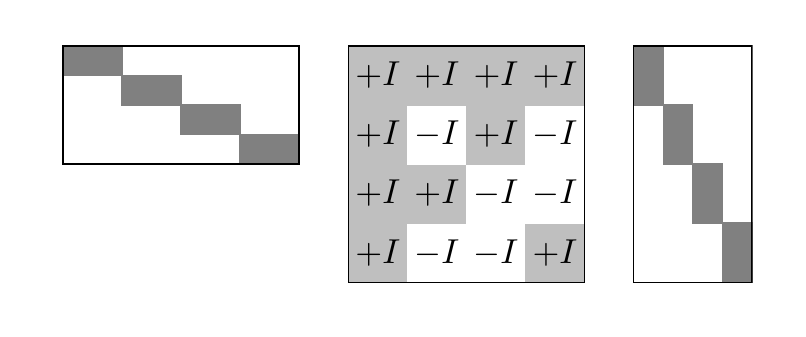} 
  \caption{The MLP is replaced by two block diagonal matrices and one block Hadamard transform.}
    \label{fig:blockdiag}
\end{minipage}
\end{figure}

Random projections can be efficiently approximated by alternating diagonal matrices and Fast Hadamard transformations using FastFood \cite{le2013fastfood}, as shown in Figure \ref{fig:fastfood}. A vector can then be projected with time complexity $\mathcal{O}(n \log n)$, significantly speeding up the original $\mathcal{O}(n^2)$ operation. Using the approach of \cite{yang2015deep}, we allow for some adaptability of the frozen layers by training the diagonal matrices. This operation dramatically reduces the number of parameters from $\mathcal{O}(n^2)$ down to $\mathcal{O}(n)$. Since in a language model, $n \sim 1000$, we approximate the cost of the structured operation to zero.

We also investigate using a combination of block diagonal matrices and block Hadamard matrices, as in Figure \ref{fig:blockdiag}. The block diagonal matrices can transform nearby features, while the block Hadamard matrix introduce long-range correlations. The parameter reduction of such a structured MLP is given by the number of blocks, and is tunable. We set the number of blocks to $4$, maintaining a relatively large number of trainable parameters in the MLP, with a cost per parameter identical to dense layers.

\subsection{Results}

\begin{figure}
\centering
\begin{minipage}[t]{0.4\textwidth}
  \includegraphics[width=\linewidth]{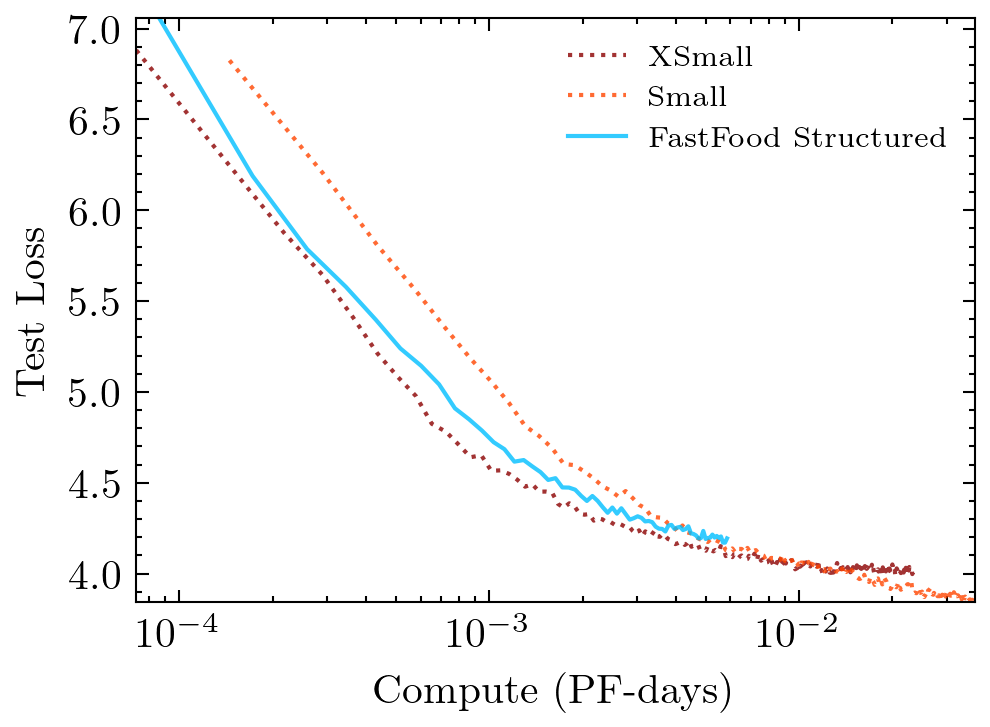} 
  \caption{Loss against compute for a model where the linear layers are replaced with adaptive FastFood.}
  \label{fig:structured_ff}
\end{minipage}
\hspace{1cm}
\begin{minipage}[t]{0.4\textwidth}
    \centering
    \includegraphics[width=\linewidth]{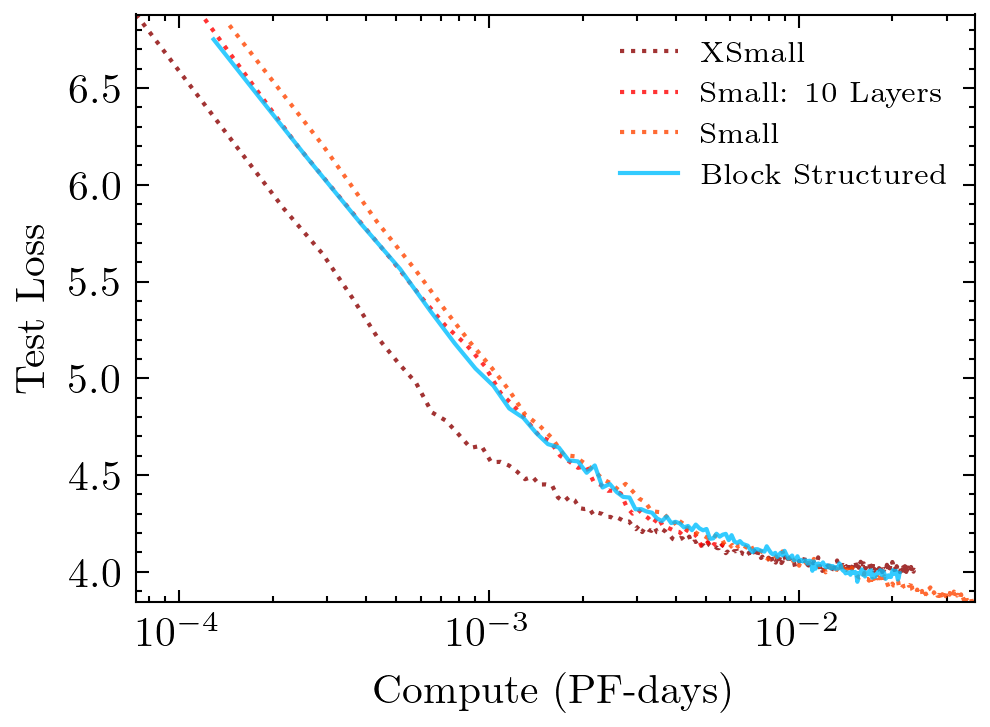} 
    \caption{Loss against compute with the MLP replaced by block diagonal matrices and block Hadamard transform.}
    \label{fig:structured_block}
\end{minipage}
\end{figure}

In Figure \ref{fig:structured_ff}, the FastFood structured model is supposed to emulate a \emph{Small} model, but actually performs like an \emph{XSmall} with a similar number of trainable parameters. Furthermore, in Figure \ref{fig:structured_block}, the behavior of the block structured model closely resembles that of a model with as many trainable parameters. Accordingly, structured models behave almost identically to fully-trainable models with similar number of trainable parameters, even though both the number of layers and the width of the model is different. The number of spurious parameters does not matter, and only the real one does. 

\section{Conclusion and outlooks}
We showed that random parameters and structured transforms do not provide an advantage over dense fully trainable layers in the loss-compute landscape of autoregressive models. Remarkably, it appears that the scaling law for these models reliably depends \emph{only} on the \emph{actual} number of trainable parameters. Unfortunately, scaling laws are not easily cheated by using spurious parameters. Our investigation points to interesting questions: is the number of trainable parameters all that actually matters? And is this true for any model architecture or specific to autoregressive language models? 

\begin{ack}
We thank Gustave Pariente and the LightOn team for useful feedback and support. This work was granted access to the HPC/AI resources of IDRIS under the allocation 2021-A0101012429 made by GENCI. Specifically, we thank the team of the Jean Zay supercomputer for their support. 

\vspace{2em}
\begin{center}
\includegraphics[width = 0.12\textwidth]{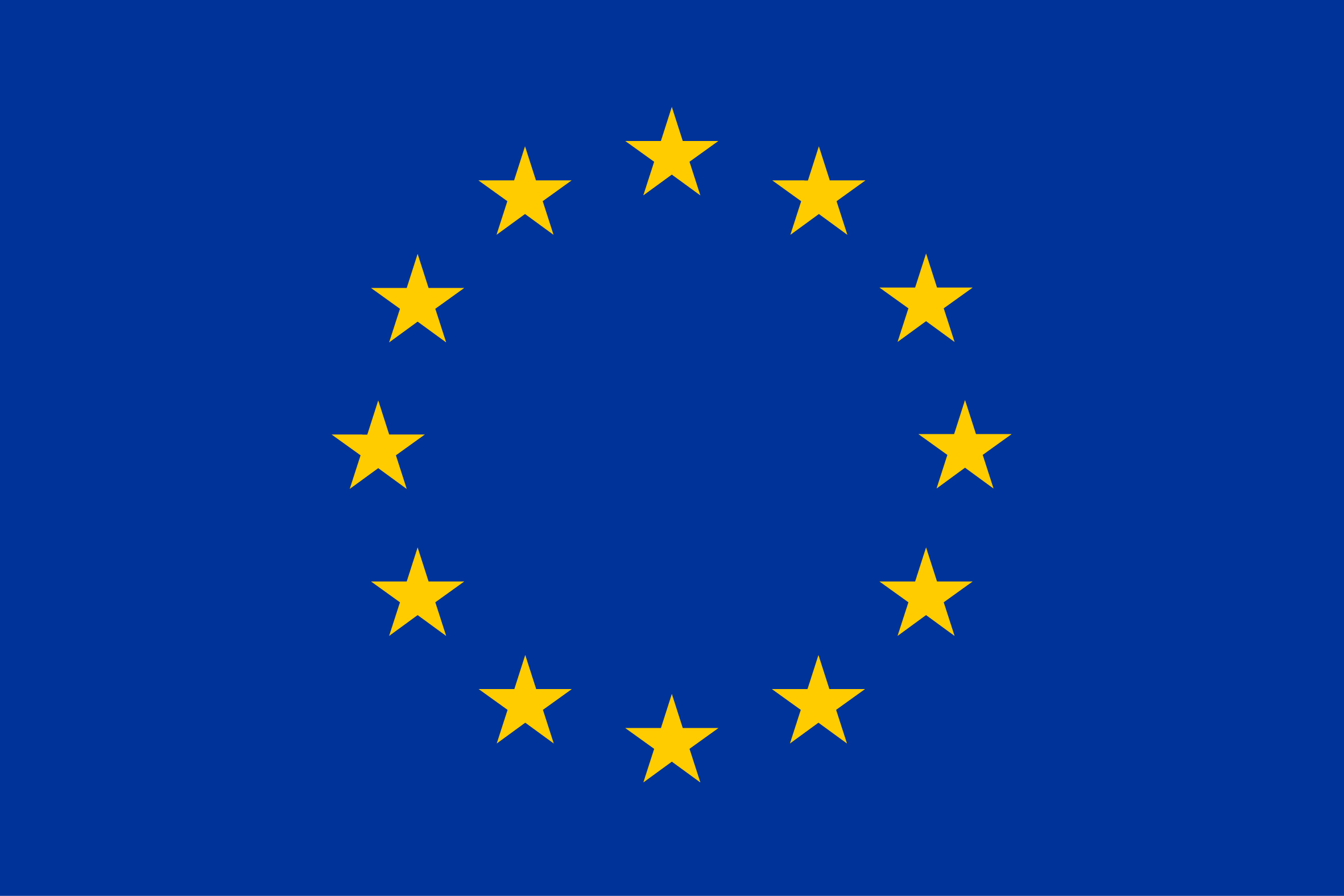}\\
This project has received funding from the European Union’s Horizon 2020 research and innovation program under the Marie Skłodowska-Curie grant agreement No 860360
\end{center}

\end{ack}

\bibliographystyle{unsrt}
\small{\bibliography{doping}
}


\clearpage
\appendix

\section{Appendix}
We include all the hyperparameter and architectural details for the baselines (Table \ref{tab:baselines}), the doped models (Table \ref{tab:doped}, and the structured models (Table \ref{tab:structured}). The doped models are built by alternating trainable transformer and frozen MLP layers, so that the first and last layers of the models are always trainable. 
\begin{table}
  \caption{Baselines}
  \label{tab:baselines}
  \centering
  \begin{tabular}{lllll}
    \toprule
    Hyperparameter     & XXSmall     & XSmall & Small & Medium \\
    \midrule
    Number of layers & $3$ & $6$ & $12$ & $24$ \\
    $d_\mathrm{model}$ & $768$ & $768$ & $768$ & $1024$ \\
    Attention heads & $12$ & $12$ & $12$ & $16$ \\
    Context size & $1024$ & $1024$ & $1024$ & $1024$ \\
    Dropout & $0.1$ & $0.1$ & $0.1$ & $0.1$ \\
    Attention dropout & $0.1$ & $0.1$ & $0.1$ & $0.1$ \\
    \midrule
    Total batch size & $80$ & $80$ & $80$ & $80$ \\
    Optimizer & Adam & Adam & Adam & Adam \\
    Peak learning rate & $5e-4$ & $5e-4$ & $5e-4$ & $3e-4$ \\
    Learning rate decay & cosine & cosine & cosine & cosine \\
    Warmup tokens & $409.6$M & $409.6$M & $409.6$M & $409.6$M \\
    Decay tokens & $32.8$G & $32.8$G & $32.8$G & $40.9$G \\
    Max training tokens & $32$G & $32$G & $32$G & $32$G \\
    Weight decay & $0.0$ & $0.0$ & $0.0$ & $0.0$ \\
    Gradient clipping & $1.0$ & $1.0$ & $1.0$ & $1.0$ \\
    \bottomrule
  \end{tabular}
\end{table}
\begin{table}
  \caption{Doped models}
  \label{tab:doped}
  \centering
  \begin{tabular}{lllll}
    \toprule
    Hyperparameter     & Doped XSmall & Doped Small & Doped Medium \\
    \midrule
    Total number of layers & $5$ & $11$ & $23$ \\
    Number of trainable layers & $3$ & $6$ & $12$ \\
    Number of frozen layers & $2$ & $5$ & $11$ \\
    $d_\mathrm{model}$ & $768$ & $768$ & $1024$ \\
    Attention heads & $12$ & $12$ & $16$ \\
    Context size & $1024$ & $1024$ & $1024$ \\
    Dropout & $0.1$ & $0.1$ & $0.1$ \\
    Attention dropout  & $0.1$ & $0.1$ & $0.1$ \\
    \midrule
    Total batch size & $80$ & $80$ & $80$ \\
    Optimizer  & Adam & Adam & Adam \\
    Peak learning rate & $5e-4$ & $5e-4$ & $3e-4$ \\
    Learning rate decay & cosine & cosine & cosine \\
    Warmup tokens & $409.6$M & $409.6$M & $409.6$M \\
    Decay tokens & $32.8$G & $32.8$G & $40.9$G \\
    Max training tokens & $32$G & $32$G & $32$G \\
    Weight decay & $0.0$ & $0.0$ & $0.0$ \\
    Gradient clipping & $1.0$ & $1.0$ & $1.0$ \\
    \bottomrule
  \end{tabular}
\end{table}

\begin{table}
  \caption{Structured Models}
  \label{tab:structured}
  \centering
  \begin{tabular}{lllll}
    \toprule
    Hyperparameter     & Structued Adaptive FastFood &  Structued Block Diagonal  \\
    \midrule
    Number of layers & $12$ & $12$ \\
    $d_\mathrm{model}$ & $1024$ & $1024$ \\
    Attention heads  & $16$ & $16$ \\
    Context size & $1024$ & $1024$ \\
    Dropout  & $0.1$ & $0.1$ \\
    Attention dropout  & $0.1$ & $0.1$ \\
    \midrule
    Total batch size& $80$ & $80$ \\
    Optimizer &  Adam & Adam \\
    Peak learning rate & $5e-4$ & $5e-4$ \\
    Learning rate decay  & cosine & cosine \\
    Warmup tokens & $409.6$M & $409.6$M \\
    Decay tokens  & $32.8$G & $40.9$G \\
    Max training tokens & $32$G & $32$G \\
    Weight decay & $0.0$ & $0.0$ \\
    Gradient clipping  & $1.0$ & $1.0$ \\
    \bottomrule
  \end{tabular}
\end{table}

\end{document}